%% file: main.tex
\documentclass[10pt,twocolumn,letterpaper]{article}

\usepackage[pagenumbers]{iccv} % To force page numbers, e.g. for an arXiv version


\definecolor{iccvblue}{rgb}{0.21,0.49,0.74}
\usepackage[pagebackref,breaklinks,colorlinks,allcolors=iccvblue]{hyperref}

\usepackage{multirow}
\usepackage{multicol}
\usepackage{{booktabs}}

\newcommand*{\affmark}[1][*]{\textsuperscript{#1}}
\newcommand{\rgbthb}{{RGB-Th-Bench}}

\def\paperID{10953} % *** Enter the Paper ID here
\def\confName{ICCV}
\def\confYear{2025}

\title{\rgbthb: A Dense benchmark for Visual-Thermal Understanding of Vision Language Models}

\author{Mehdi Moshtaghi\affmark[1,2,3] \quad Siavash H. Khajavi\affmark[1,3]\quad Joni Pajarinen\affmark[1]\\
{\affmark[1]Aalto University\quad\quad\affmark[2]KTH Royal Institute of Technology\quad\quad\affmark[3]Detectium Oy} %\vspace{-1.5em}
}

\begin{document}
\maketitle
\input{sec/0_abstract}    
\input{sec/1_intro}
\input{sec/2_related_work}
\input{sec/3_rgb_th_bench}
\input{sec/4_experiments}
\input{sec/5_conclusion}

{
    \small
    \bibliographystyle{ieeenat_fullname}
    % \bibliography{main}
    
}

\input{sec/X_suppl}

\end{document}

%% file: sec/0_abstract.tex
\begin{abstract}
We introduce \textbf{RGB-Th-Bench}, the first benchmark designed to evaluate the ability of Vision-Language Models (VLMs) to comprehend RGB-Thermal image pairs. While VLMs have demonstrated remarkable progress in visual reasoning and multimodal understanding, their evaluation has been predominantly limited to RGB-based benchmarks, leaving a critical gap in assessing their capabilities in infrared vision tasks. Existing visible-infrared datasets are either task-specific or lack high-quality annotations necessary for rigorous model evaluation. To address these limitations, RGB-Th-Bench provides a comprehensive evaluation framework covering 14 distinct skill dimensions, with a total of 1,600+ expert-annotated Yes/No questions. The benchmark employs two accuracy metrics: a standard question-level accuracy and a stricter skill-level accuracy, which evaluates model robustness across multiple questions within each skill dimension. This design ensures a thorough assessment of model performance, including resilience to adversarial and hallucinated responses. We conduct extensive evaluations on 19 state-of-the-art VLMs, revealing significant performance gaps in RGB-Thermal understanding. Our results show that even the strongest models struggle with thermal image comprehension, with performance heavily constrained by their RGB-based capabilities. Additionally, the lack of large-scale application-specific and expert-annotated thermal-caption-pair datasets in pre-training is an important reason of the observed performance gap. RGB-Th-Bench highlights the urgent need for further advancements in multimodal learning to bridge the gap between visible and thermal image understanding. The dataset is available through \href{https://drive.google.com/file/d/1aNaKYpI0lirhDUm34eq6PChA0pmjoXUD/view?usp=sharing}{this link}, and the evaluation code will also be made publicly available.
\end{abstract}

%% file: sec/1_intro.tex
\section{Introduction}
\label{sec:intro}

%%%%%%%%%%%%%%%%%%%%%%%%%%%%%%%%%%%%%%%%%%%%%%%%
\begin{figure*}[h]
  \centering
  \includegraphics[width=\linewidth]{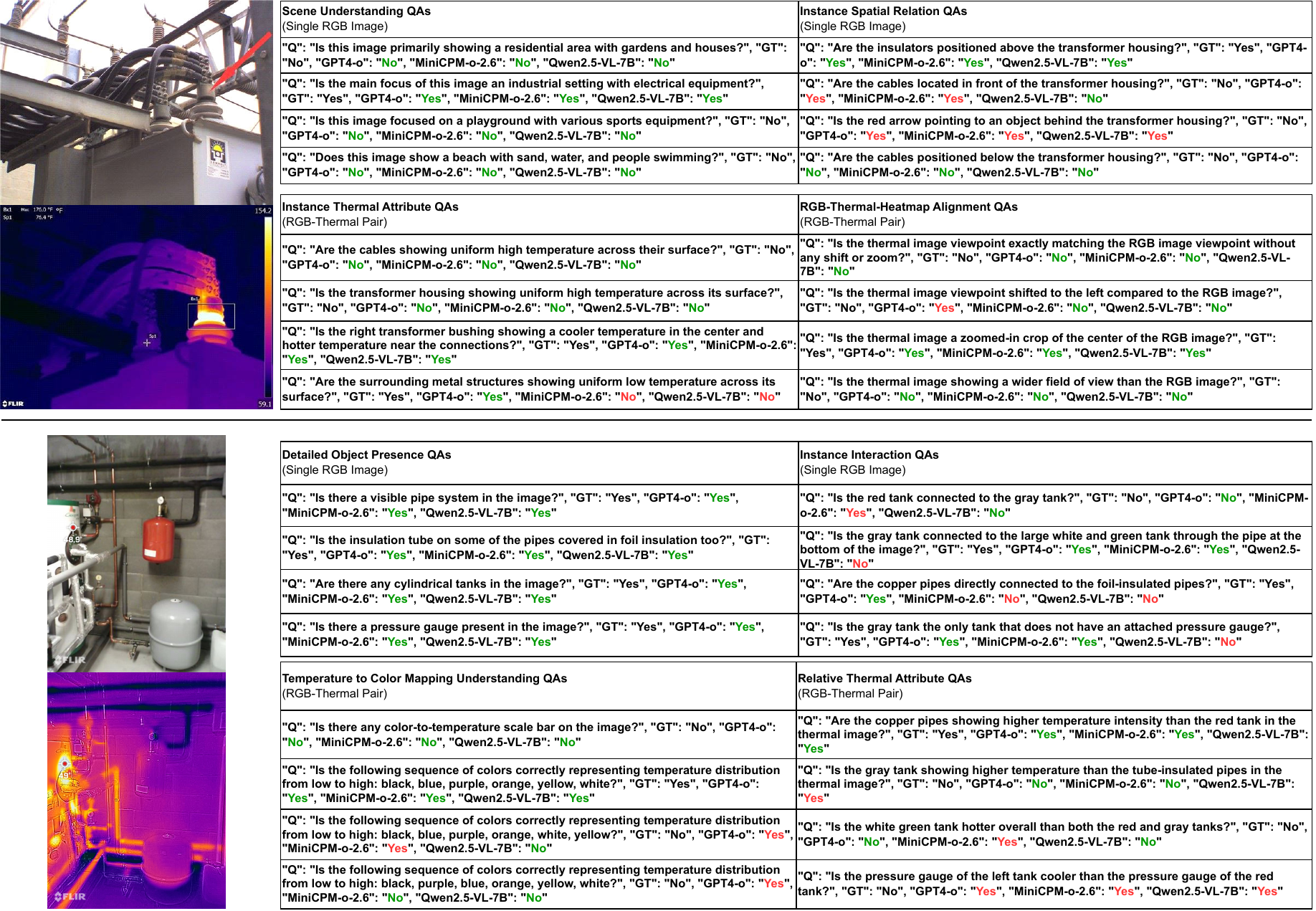}
  \caption{Two data samples from RGB-Th-Bench. Due to space limit, here we show only a subset of 16 Q/As across 4 skill dimensions for each sample, and the response from 3 selected VLMs.}
  \label{fig:benchmark_samples}
\end{figure*}
%%%%%%%%%%%%%%%%%%%%%%%%%%%%%%%%%%%%%%%%%%%%%%%%

Vision-Language Models (VLMs)~\cite{qwenvl_bai2023, Qwen2VL, 2024-minicpm-v, chen2024internvl1.5, zhang2024internlmxcomposer25, llava_liu2024, llavanext_liu2024, tested_models_2024_llavaonevision, tested_models-2024-pixtral12b, chameleon2024, claude_sonnet, gemini_pro_reid2024} have made remarkable progress in recent years, leveraging large-scale image-text datasets to align visual representations with the powerful reasoning capabilities of Large Language Models (LLMs)~\cite{openai2024gpt4technicalreport,LLaMA_Touvron2023, dubey2024llama3, qwen_bai2023, cai2024internlm2}. To systematically evaluate their multimodal understanding, numerous benchmarks have been developed, such as MME~\cite{visling_mme_Fu2023}, MMBench~\cite{visling_MMBench_Liu2023}, SEED-Bench~\cite{visling_seedbench_li2023}, and MM-Vet~\cite{visling_mmvet_yu2023}. These benchmarks have become widely recognized for assessing various vision-language skill dimensions, including core perception tasks, domain knowledge, reasoning abilities, and application-specific tasks.

Despite the success of existing VLM benchmarks, they primarily focus on RGB images and lack comprehensive evaluation for multimodal tasks involving infrared (thermal) imagery. Visible-infrared paired datasets are rare and often designed for specific applications, such as pedestrian detection, autonomous driving, military applications, or object tracking~\cite{OTCBVS, cvc13, cvc14, cvc15, tno, ino_video_nodate, dataset_kaist_pedestrian, dataset_llvip_rgbthermalpair_pedestrian, flir}. Moreover, a critical gap is the absence of high-quality, expert-annotated datasets with aligned visible-thermal captions, which are essential for training and evaluating VLMs on thermal image understanding. As a result, the evaluation of existing VLMs on RGB-Thermal paired data is an unexplored area in vision-language research.

To this end, we introduce \textbf{RGB-Th-Bench}, the \textbf{first} benchmark designed to rigorously evaluate the ability of VLMs to comprehend RGB-Thermal image pairs.

RGB-Th-Bench currently comprises over 1,600 Yes/No questions across 14 skill dimensions. Each skill dimension includes four questions per image, resulting in 116 questions per skill dimension and 56 questions per RGB image in total. This design ensures that RGB-Th-Bench is both dense and comprehensive.

The benchmark employs two accuracy metrics: a question-level accuracy metric, where random performance is 50\%, and a stricter skill-level accuracy metric, where all four questions within a skill dimension must be answered correctly for a "pass". This dual-metric approach allows for a nuanced evaluation of model's performance including its robustness against hallucinations and adversarial questions.

To evaluate RGB-Th-Bench, we conduct extensive experiments on a diverse set of 19 VLMs, including both open-source and closed-source models, reporting their performance across various skill dimensions and metrics. Our results reveal significant performance gaps in RGB-Thermal understanding. Notably, we find that most models struggle with RGB-Thermal tasks, and their performance is ultimately upper-bounded by their core RGB-based capabilities.

Our benchmark highlights the need for substantial improvements in VLMs before they can be effectively deployed in real-world RGB-Thermal applications. We hope RGB-Th-Bench serves as a foundational resource for advancing multimodal research and improving vision-language models in the challenging domain of thermal image understanding.

Inspired by "Machine Learning that Matters"~\cite{wagstaff2012ml_matters}, we believe that designing impactful benchmarks and metrics can drive upstream research efforts, guiding the selection of training datasets, structuring of experiments, and definition of objective functions. To this end, this work demonstrates how improvements in VLM accuracy can translate into meaningful impact for real-world problem domains.

%% file: sec/2_related_work.tex
\section{Related Work}
\label{sec:2_related_work}

%%%%%%%%%%%%%%%%%%%%%%%%%%%%%%%%%%%%%%%%%%%%%%%%%%%%%%%%%
\begin{table*}[h]
\centering
\footnotesize
\begin{tabular}{lcccccc}
\toprule
\textbf{Dataset} & \textbf{\#Thermal} & \textbf{RGB-Th Pairs} & \textbf{Text Ann.} & \textbf{Task} & \textbf{Obj. Ctg.} & \textbf{Features} \\
\midrule
OTCBVS~\cite{OTCBVS} & $\sim$16 sets & No & No & Ped Det, Wpn Det & Varies & Task-specific, fragmented \\
CVC-13~\cite{cvc13} & 23 & Yes & No & Ped Det & Peds & Small size, urban focus \\
CVC-14~\cite{cvc14} & N/A & Yes & No & Ped Det & Peds & Misaligned pairs, no fusion \\
CVC-15~\cite{cvc15} & 100 & Yes & No & Ped Det & Peds & BB ann. provided \\
KAIST~\cite{dataset_kaist_pedestrian} & 95k+ & Yes & No & Self-Driving, Ped Det & Ped, Veh & Well-aligned, day/night \\
LLVIP~\cite{dataset_llvip_rgbthermalpair_pedestrian} & 15,488 & Yes & No & LL Ped Det & Peds & High-quality alignment \\
INO~\cite{ino_video_nodate} & N/A (video) & Yes & No & Obj Det, Ped Track & Peds & Weather-varied, video \\
FLIR~\cite{flir} & 26,442 & No & No & Self-Driving, Obj Det & 15 Ctg. & BB ann. provided \\
TNO~\cite{tno} & 63 & Yes & No & Img Fus & Mil objs & Multispectral (Vis, NIR, LWIR) \\
27UAV~\cite{27UAV} & 27 & Yes & No & UAV Surv & N/A & Small scale, UAV focus \\
\bottomrule
\end{tabular}
\caption{Comparison of thermal image datasets. Abbreviations: \textbf{\#Thermal} = Number of thermal images; \textbf{RGB-T Pairs} = Paired RGB-thermal images; \textbf{Text Ann.} = Text or caption annotations; \textbf{Task} = Task focus (\textbf{Det}: Detection, \textbf{Ped}: Pedestrian, \textbf{Wpn}: Weapon, \textbf{LL}: Low-light, \textbf{Obj}: Object, \textbf{Track}: Tracking, \textbf{Img Fus}: Image Fusion, \textbf{UAV Surv}: UAV-based Surveillance); \textbf{Obj. Ctg.} = Object categories (\textbf{Veh}: Vehicles, \textbf{Mil objs}: Military objects, \textbf{15 Ctg.}: 15 annotated categories); \textbf{Features}: Unique dataset features (\textbf{BB ann.}: Bounding box annotations, \textbf{Vis, NIR, LWIR}: Visible, near-infrared, longwave-infrared data).}
\label{tab:thermal_datasets}
\end{table*}
\normalsize
%%%%%%%%%%%%%%%%%%%%%%%%%%%%%%%%%%%%%%%%%%%%%%%%%%%%%%%%%

\textbf{Vision-Language Models.}  
Building on the remarkable success of LLMs~\cite{openai2024gpt4technicalreport,LLaMA_Touvron2023, dubey2024llama3, qwen_bai2023, cai2024internlm2}, recent research has focused on VLMs~\cite{qwenvl_bai2023, Qwen2VL, 2024-minicpm-v, chen2024internvl1.5, zhang2024internlmxcomposer25, llava_liu2024, llavanext_liu2024, tested_models_2024_llavaonevision, tested_models-2024-pixtral12b, chameleon2024} to enhance multimodal understanding. These models achieve this by aligning the visual features extracted from pre-trained image encoders with LLMs using large-scale image-text datasets.

\textbf{Recent Benchmarks for Vision-Language Models.}  
To systematically assess the capabilities of VLMs, several benchmarks have been introduced, such as MME~\cite{visling_mme_Fu2023}, MMBench~\cite{visling_MMBench_Liu2023}, SEED-Bench~\cite{visling_seedbench_li2023}, MMMU~\cite{visling_MMMU_Yue2023}, MM-Vet~\cite{visling_mmvet_yu2023}, and NaturalBench~\cite{naturalbench}. These benchmarks have become widely recognized as standard evaluation tools, as reflected in their prominence on performance leaderboards~\cite{2023opencompass}.  

Despite their broad adoption, these benchmarks primarily assess single-image text comprehension and lack support for tasks involving multiple images or interleaved image-text sequences. While this was a reasonable limitation in earlier VLMs designed for single-image inputs, recent advancements~\cite{zhang2024internlmxcomposer25, Qwen2VL, 2024-minicpm-v, openai2024gpt4technicalreport, gemini_pro_reid2024} have enabled models to process multiple images, sequential video frames, and arbitrarily interleaved multimodal inputs. In response to this shift, new benchmarks such as MM-Vet2~\cite{yu2024mmvet2}, SeedBench2~\cite{visling_seedbench2_li2023}, and Seed-Bench2-Plus~\cite{visling_seedbench2_plus_li2024} have been introduced as extensions to their predecessors.  

Most VLM benchmarks evaluate model performance across multiple skill dimensions, which can be broadly categorized into four overlapping domains:  
\begin{itemize}  
    \item \textbf{Core perception skills}, including scene understanding, object existence, object attributes, and spatial reasoning~\cite{visling_mme_Fu2023, visling_mmvet_yu2023, visling_seedbench_li2023, visling_MMBench_Liu2023, hallucination_POPE_Li2023EvaluatingOH}.  
    
    \item \textbf{Domain knowledge}, encompassing both general knowledge and scientific understanding~\cite{visling_science_qa_Lu2022, visling_MMMU_Yue2023, visling_mme_Fu2023, visling_seedbench2_li2023}.  
    
    \item \textbf{Reasoning abilities}, including mathematical reasoning, logical reasoning, commonsense reasoning, and attribute-based reasoning~\cite{visling_mme_Fu2023, visling_MMMU_Yue2023, ComparativeReasoningBenchmark, wang2025enigmaevalbenchmarklongmultimodal}.  
    
    \item \textbf{Application-specific skills}, such as celebrity recognition~\cite{visling_mme_Fu2023, visling_MMBench_Liu2023}, meme comprehension~\cite{visling_seedbench2_li2023}, landmark recognition~\cite{visling_seedbench2_li2023}, chart interpretation~\cite{visling_seedbench2_plus_li2024}, map reading~\cite{visling_seedbench2_plus_li2024}, OCR~\cite{visling_mmvet_yu2023}, GUI navigation~\cite{lin2024videogui-bench}, and screenshot-to-code translation~\cite{si2024design2code}.  
\end{itemize}  

\textbf{Visible-Infrared Image Pair Datasets.}
Infrared image datasets are considerably rarer than their visible-light counterparts, with most existing datasets designed for specific tasks such as pedestrian detection, object tracking, and image fusion. \Cref{tab:thermal_datasets} presents the comparison among these datasets. Notably, datasets like OTCBVS~\cite{OTCBVS}, and and FLIR~\cite{flir} only provide Thermal images, which limits their use for multimodal learning. CVC-13/14/15~\cite{cvc13, cvc14, cvc15}, 27UAV~\cite{27UAV}, and TNO~\cite{tno} are small in scale, containing less than 100 RGB-Thermal pairs of highly similar scenes. Although KAIST~\cite{dataset_kaist_pedestrian}, and LLVIP~\cite{dataset_llvip_rgbthermalpair_pedestrian} provide larger collections, they remain task-specific with limited object diversity, limiting their broader applicability in vision-language research.

Despite these efforts, no existing dataset provides captioned infrared images or visual instruction annotations, which are essential for advancing multimodal understanding of thermal imagery. Additionally, key applications such as heat and energy loss detection remain underrepresented due to annotation challenges, as infrared images often lack well-defined object boundaries. The absence of such datasets highlights the need for a more comprehensive benchmark that integrates textual descriptions with aligned visible-infrared image pairs, addressing a critical gap in vision-language model evaluation.

%% file: sec/3_rgb_th_bench.tex
\section{RGB-Thermal Image Benchmark}
\label{sec:3_rgb_th_bench}

In this work, we developed a high-quality dense multimodal benchmark for visual-thermal understanding, called \rgbthb, that requires comprehensive skills or capabilities to stimulate the research and provide a foundation for future advancements in infrared image analysis and visual-thermal multimodal research.

\subsection{Data Collection}
\begin{figure}[t]
  \centering
  \includegraphics[width=\linewidth]{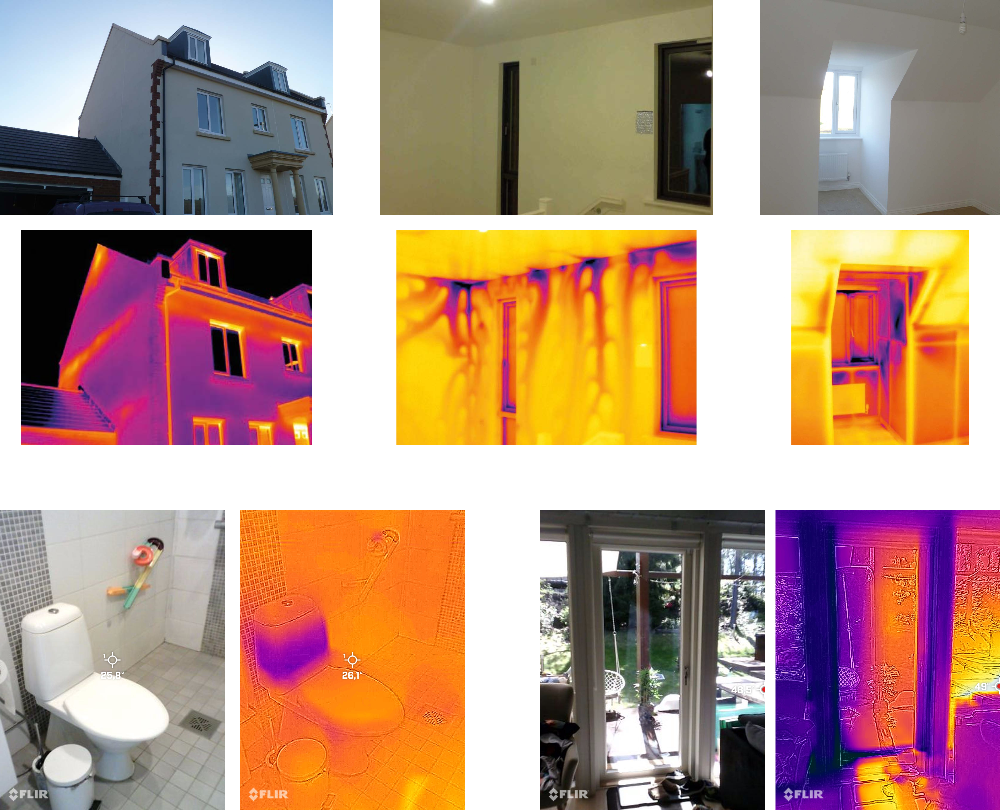}
  \caption{Some RGB-thermal pair samples in residential settings}
  \label{fig:resident_pair_img_exmp}
\end{figure}

\begin{figure}[t]
  \centering
  \includegraphics[width=\linewidth]{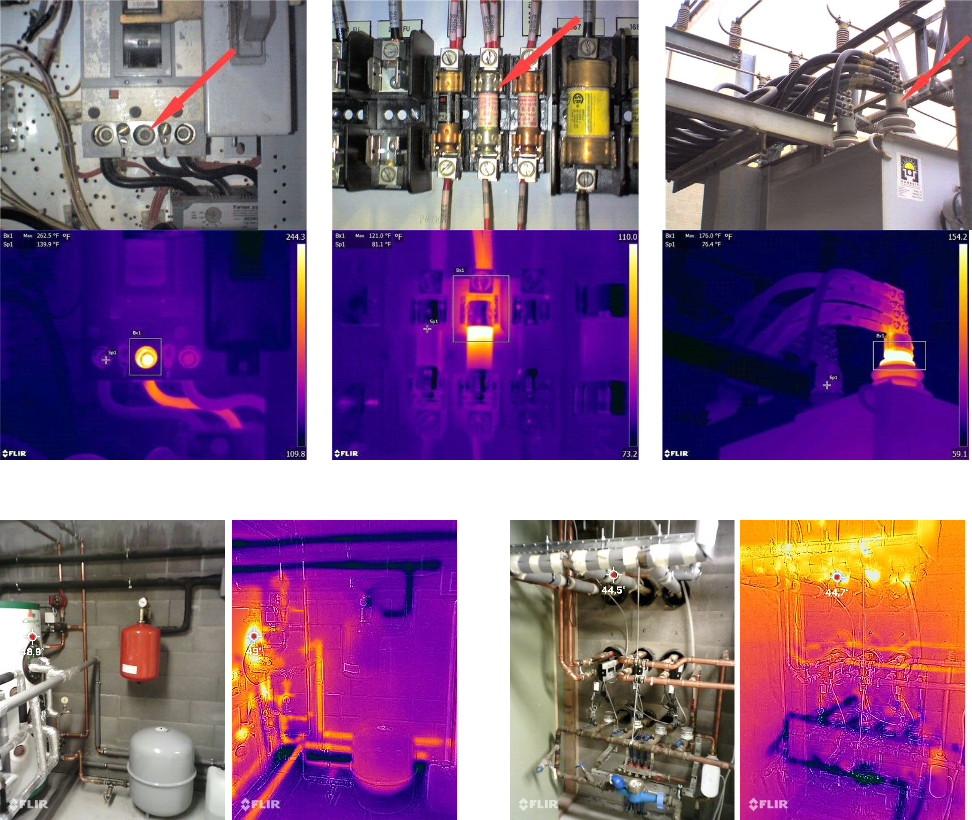}
  \caption{Some RGB-thermal pair samples in industrial settings}
  \label{fig:indust_pair_img_exmp}
\end{figure}

The benchmark includes a total of 1624 questions and 58 images (29 RGB-Thermal sample pairs). Similar to MME~\cite{visling_mme_Fu2023}, POPE~\cite{hallucination_POPE_Li2023EvaluatingOH}, and NatrualBench~\cite{naturalbench}, these questions entail Yes/No responses.
We do not use any image or question from any other dataset or benchmark. The images are either manually collected from internal documents or directly taken using "FLIR ONE Edge Pro" camera~\cite{FLIR_ONE_Edge_Pro}. All the questions and their ground truths answers are manually annotated by us, under the supervision of domain experts. As shown in \Cref{fig:resident_pair_img_exmp} and \Cref{fig:indust_pair_img_exmp}, most of the scenes are of importance for risk and anomaly detection in industrial or residential settings for example for detection of poor insulation, gaps and cracks, leakage, overheating, electrical issue, and fire. Intentionally, this benchmark is not designed for military or surveillance applications.

\subsection{Evaluation Dimensions}
\label{sec:eval_dim}
For each sample pair, there are a total of 56 questions across 14 core evaluation or skill dimensions, resulting in four independent Yes/No questions per skill dimension for each sample. This way, we make sure that every sample pair is thoroughly tested from all possible skill dimensions to make our benchmark a dense evaluation setting, which is unlike the benchmarks that ask only one or two questions from each sample image, like MME~\cite{visling_mme_Fu2023}, MM-Vet~\cite{visling_mmvet_yu2023} and Seed-Bench2~\cite{visling_seedbench2_li2023}.
These skills are grouped into two types of prompt:
\begin{itemize}
    \item \textbf{Single RGB \& Text (RGB-Txt):} This prompt-group is necessary to evaluate model performance without thermal input, so that we can control the effect of addition of the thermal image to the same input. This prompt-group consists of seven diverse evaluation dimensions including Scene Understanding (Scene), Detailed Object Presence (ObjPr), Instance Attributes (Attr), Instance Location w.r.t. Image (LocIm), Instance Spatial Relation (SpRel), Instance Counting (Count), and Instance Interaction (Interact).

    \item \textbf{RGB-Thermal pair \& Text (RGB-Th-Txt):} This prompt-group is to evaluate model performance with thermal input, and similarly includes seven diverse evaluation dimensions, including Temperature to Color Mapping Understanding (Temp2Col), RGB-Thermal-Heatmap Alignment (HeatAlign), Instance Thermal Attribute (ThermAttr), Relative Thermal Attribute (RelTherm), Warmest Areas Detection (Warmest), Coldest Areas Detection (Coldest), and Risk and Anomaly Detection (Anomaly).
\end{itemize}
The detailed guideline for annotations of all evaluation dimensions is provided in the supplementary materials.% in \Cref{sec:guidelines}.

%%%%%%%%%%%%%%%%%%%%%%%%%%%%%%%%%%%%%%%%%%%%%%%%%%%%%%%%%%%
\begin{table*}[h]
\centering
\scriptsize
\begin{tabular}{lccccccc}
\toprule
\textbf{Benchmarks} & \textbf{\#Skills}& \textbf{Visual Modality} & \textbf{Visual Source} & \textbf{\#Q/A Annotations} & \textbf{\#Q/A Source} & \textbf{Answer Type} & \textbf{Question Density} \\
\midrule
MMBench~\cite{visling_MMBench_Liu2023} & 20 & RGB(s) & mixed & 3217 & human & free-form & 1 \\

HallusionBench~\cite{guan2024hallusionbench} & 9 & RGB & web & 1129 & human & Y/N & 3.26 \\

MMVet~\cite{visling_mmvet_yu2023} & 6 & RGB & mixed & 218 & human & free-form & 1.09 \\

POPE~\cite{hallucination_POPE_Li2023EvaluatingOH} & 1 & RGB & MSCOCO~\cite{ObjDet_MSCOCO_Lin2014} & 3000 & mixed & Y/N & 6 \\

SEEDBench2~\cite{visling_seedbench2_li2023} & 27 & RGB(s) \& video & other datasets & \textbf{24371} & human\&GPT & A/B/C/D & 1 \\

LLaVA-Bench~\cite{llava_liu2024} & 3 & RGB & web & 64 & human & free-form & 2.5 \\

MME~\cite{visling_mme_Fu2023} & 14 & RGB & mixed & 2374 & human & Y/N & 2 \\

\rgbthb (Ours) & 14 & RGB-Thermal & \textbf{self-sourced} & 1624 & human & Y/N & \textbf{56} \\
\bottomrule
\end{tabular}
\caption{Comparison of RGB-Th-Bench with some other established benchmarks. "\#Skills" and "Question Density" mean the number of skill dimensions and the number of questions per each visual sample, respectively.}
\label{tab:benchmarks comparison}
\end{table*}

\normalsize
%%%%%%%%%%%%%%%%%%%%%%%%%%%%%%%%%%%%%%%%%%%%%%%%%%%%%%%%%%%

\subsection{Instruction Design}
To enable the collection of quantitative performance statistics, the instruction design is structured to elicit "Yes" or "No" responses from the model. Consequently, the instruction comprises two components: a concise question and a fixed context. The context for all questions in the "RGB-Txt" prompt-group is \textit{Based on this image, answer the following question with strictly either "Yes" or "No", without any extra explanation}, and the one for questions in the "RGB-Th-Txt" prompt-group is \textit{Based on these two images, and the fact that the second image is the thermal image taken from the same scene as the first image, answer the following question with strictly either "Yes" or "No", without any extra explanation.}

If the model’s response does not include Yes or No (case insensitive), it is marked as a disregard to the simple Yes/No instruction and considered as a “Fail”. Note that we do not require strict exact matching, for example the model can respond like "The answer is yes" and we qualify it as a "Yes", since it includes either case-insensitive "Yes" or "No". Similar to MME~\cite{visling_mme_Fu2023}, we argue that a good VLM (especially after instruction tuning) should be able to follow such a simple instruction, which is also very common in everyday life.

\subsection{Evaluation Strategy}
Since the output of the model is limited to two types (“Yes” or “No”), it is convenient to measure the accuracy metric on both question-level and skill-level for each skill dimension:
\begin{itemize}
    \item Question-level Accuracy (QAcc): The percentage of the questions answered correctly, with random baseline score of 50\%.
    \item Skill-level Accuracy (SAcc): if the model correctly answers all the four questions of a skill dimension of a sample pair, then we consider that as a “Pass” for that skill, otherwise a “Fail”.  The random baseline score is 6.25\% in this case.
\end{itemize}
It can be seen that Skill-level Accuracy is a stricter measurement but also better reflects the comprehensive understanding degree of the model in its respective skill. In addition, we calculate the overall score of each prompt-group based on the average of accuracy score in their skill dimension.

\subsection{Advantages of RGB-Th-Bench}
\Cref{tab:benchmarks comparison} provides a comparison between our work and some other established VLM benchmarks. 

RGB-Th-Bench is  \textbf{the first and the only} benchmark that focuses on both RGB and thermal understanding, and also introduces several key advantages that make it a robust tool for advancing research in visual-thermal understanding and infrared image analysis. Below, we outline the primary benefits:

\begin{itemize}
    \item \textbf{Flexibility and Robust Metrics:} RGB-Th-Bench provides flexibility in evaluation by allowing for the inclusion of multiple questions per evaluation dimension. If the scene lacks sufficient number of objects to generate diverse questions, multiple questions can focus on specific objects, often incorporating challenging or adversarial cases, such as negative questions. This ensures a more robust assessment of a model’s resilience to adversarial inputs and susceptibility to hallucination.

    \item \textbf{Fairness in Evaluation:} Unlike multiple-choice question-answering benchmarks, where the model benefits from having all answer options within its context and knowing implicitly that only one option is correct, RGB-Th-Bench uses independent questions. Each query stands alone, requiring the model to analyze and respond without reliance on contextual cues from a predefined set of options. This independence ensures a more equitable and unbiased evaluation of the model’s capabilities.

    \item \textbf{Dense and Comprehensive Testing:} RGB-Th-Bench rigorously evaluates models across 14 core aspects of visual-thermal understanding by carefully crafting 56 questions for each RGB-Thermal pair sample. Instead of posing a single or limited set of questions per image, our benchmark tests the model comprehensively, examining its robustness and generalization across multiple dimensions. This dense approach ensures a thorough assessment of the model’s strengths and weaknesses, promoting a deeper understanding of its performance.

    \item \textbf{High-Quality Annotations:} All questions and answers in RGB-Th-Bench are manually annotated by us with the supervision of domain experts, ensuring their relevance and accuracy. Additionally, the images used in the benchmark are sourced exclusively from internal documents or directly taken by us, and not from other datasets or from the web, guaranteeing that they are not included in pre-training or fine-tuning datasets. This ensures a true test of the model’s generalization capability and eliminates potential biases arising from prior exposure to the data.
\end{itemize}

% /////////////////////////////////////

%% file: sec/4_experiments.tex
\section{Experiments}
\label{sec:4_experiments}
\subsection{Target Models}
We evaluate a total of 19 recent VLMs that support multi-image inputs, including Chameleon-7B~\cite{chameleon2024}, InternVL2.5 series~\cite{tested_models_2025_internvl2.5}, SmolVLM~\cite{tested_models_2025_smolvlm}, Llava-Next series~\cite{liu-2024-llava-next_mistral_vicuna, li-2024-llava-next-llama3, tested_models_li-2024-llava-next-interleave}, Pixtral-12B~\cite{tested_models-2024-pixtral12b}, Llava-Onevision-ov series~\cite{li2024llavaonevision}, MiniCPM-V-2.6~\cite{2024-minicpm-v}, MiniCPM-o-2.6~\cite{OpenBMB-MiniCPMo-2025}, Qwen2-VL~\cite{Qwen2VL} and Qwen2.5-VL~\cite{tested_models_2025_Qwen25VL} series, and GPT4o~\cite{openai2024gpt4technicalreport}. To facilitate the evaluation process, we incorporate VLMEvalKit~\cite{duan2024vlmevalkit} as our evaluation toolkit.

%%%%%%%%%%%%%%%%%%%%%%%%%%%%%%%%%%%%%%%%%%%%%%%%
\begin{table*}[h]
\centering
\footnotesize % Set font size to footnotesize
\begin{tabular}{@{}l|c|c|cc|cc|cc@{}}
\toprule

Models & Language Model & Vision Model & \multicolumn{2}{c}{RGB-Txt} & \multicolumn{2}{c}{RGB-Th-Txt} & \multicolumn{2}{c}{Overall}\\
\cmidrule(lr){4-5} \cmidrule(lr){6-7} \cmidrule(lr){8-9}
& & & QAcc & SAcc & QAcc & SAcc & QAcc & SAcc \\

\midrule
\multirow{1}{*}{Random Baseline} & - & - & 50 & 6.25 & 50 & 6.25 & 50 & 6.25 \\
\multirow{1}{*}{Statistically Significant} & - & - & $>$53 & $>$8 & $>$53 & $>$8 & $>$52 & $>$7.5 \\

\midrule
\multirow{1}{*}{Chameleon-7B~\cite{chameleon2024}} & Unified-Mixed-Modal & Unified-Mixed-Modal & 26.48 & 0.49 & 31.54 & 0.49 & 29.01 & 0.49 \\

\multirow{1}{*}{InternVL2.5-1B~\cite{tested_models_2025_internvl2.5}} & Qwen2.5-0.5B & InternVit-300M-v2.5 & 65.89 & 23.15 & 56.68 & 4.93 & 61.28 & 14.04 \\

\multirow{1}{*}{InternVL2.5-2B~\cite{tested_models_2025_internvl2.5}} & InternLM2.5-1.8B & InternVit-300M-v2.5 & 67.86 & 25.62 & 61.60 & 5.42 & 64.73 & 15.52 \\

\multirow{1}{*}{Llava-Onevision-1B-ov~\cite{li2024llavaonevision}} & Qwen2-0.5B & SigLip-400M & 69.58 & 30.05 & 47.66 & 4.93 & 58.62 & 17.49 \\

\multirow{1}{*}{SmolVLM-2.3B~\cite{tested_models_2025_smolvlm}} & SmolLM2-1.7B & SigLip-400M & 69.09 & 28.57 & 53.45 & 9.85 & 61.27 & 19.21 \\

\multirow{1}{*}{Qwen2-VL-2B-Instruct~\cite{Qwen2VL}} & Qwen2-1.5B & QwenViT & 75.25 & 35.96 & 58.64 & 4.93 & 66.95 & 20.44 \\

\multirow{1}{*}{Llava-Next-Mistral-7.6B~\cite{liu-2024-llava-next_mistral_vicuna}} & Mistral-7B & CLIP-ViT-L/14 & 77.83 & 39.41 & 52.82 & 3.45 & 65.33 & 21.43 \\

\multirow{1}{*}{Llava-Next-Interleave-8B~\cite{tested_models_li-2024-llava-next-interleave}} & Qwen1.5-7B & SigLip-400M & 76.72 & 40.39 & 47.30 & 4.93 & 62.01 & 22.66 \\

\multirow{1}{*}{Llava-Next-Llama3-8B~\cite{li-2024-llava-next-llama3}} & Llama3-8B-Instruct & CLIP-ViT-L/14 & 77.22 & 38.42 & 58.63 & \colorbox{SkyBlue}{10.34} & 67.92 & 24.38 \\

\multirow{1}{*}{Pixtral-12.4B-2409~\cite{tested_models-2024-pixtral12b}} & Mistral-Nemo-12B & Pixtral-ViT-400M & 76.85 & 39.41 & 59.13 & 9.36 & 67.99 & 24.38 \\

\multirow{1}{*}{Qwen2.5-VL-3B-Instruct~\cite{tested_models_2025_Qwen25VL}} & Qwen2.5-7B & QwenViT & \colorbox{Yellow}{80.17} & \colorbox{YellowGreen}{46.80} & 53.68 & 5.42 & 66.93 & 26.11 \\

\multirow{1}{*}{Llava-Onevision-8B-ov~\cite{li2024llavaonevision}} & Qwen2-7B & SigLip-400M & 79.19 & 42.86 & 60.73 & \colorbox{Yellow}{10.84} & 69.96 & 26.85 \\

\multirow{1}{*}{InternVL2.5-4B~\cite{tested_models_2025_internvl2.5}} & Qwen2.5-3B & InternViT-300M-v2.5 & 79.31 & 43.84 & 59.13 & 9.85 & 69.22 & 26.85 \\

\multirow{1}{*}{InternVL2.5-8B~\cite{tested_models_2025_internvl2.5}} & InternLM2.5-7B & InternViT-300M-v2.5 & 79.68 & \colorbox{Yellow}{46.31} & \colorbox{SkyBlue}{63.08} & 8.37 & 71.38 & 27.34 \\

\multirow{1}{*}{Qwen2.5-VL-7B-Instruct~\cite{tested_models_2025_Qwen25VL}} & Qwen2.5-7B & QwenViT & 75.62 & 37.44 & \colorbox{YellowGreen}{\textbf{67.26}} & \colorbox{YellowGreen}{\textbf{17.24}} & \colorbox{SkyBlue}{71.44} & 27.34 \\

\multirow{1}{*}{MiniCPM-V-2.6-8B~\cite{2024-minicpm-v}} & Qwen2-7B & SigLip-400M & \colorbox{SkyBlue}{79.80} & \colorbox{SkyBlue}{45.32} & \colorbox{Yellow}{63.31} & 9.85 & \colorbox{Yellow}{71.56} & \colorbox{SkyBlue}{27.59} \\

\multirow{1}{*}{MiniCPM-o-2.6-8B~\cite{OpenBMB-MiniCPMo-2025}} & Qwen2.5-7B & SigLip-400M & \colorbox{YellowGreen}{81.16} & \colorbox{Yellow}{46.31} & 62.20 & 9.85 & \colorbox{YellowGreen}{71.68} & \colorbox{Yellow}{28.08} \\

\multirow{1}{*}{Qwen2-VL-7B-Instruct~\cite{Qwen2VL}} & Qwen2-7B & QwenViT & \colorbox{YellowGreen}{81.16} & \colorbox{YellowGreen}{46.80} & 60.61 & 9.85 & 70.88 & \colorbox{YellowGreen}{28.33} \\
\midrule
\multirow{1}{*}{GPT4o-20241120~\cite{openai2024gpt4technicalreport}} & closed-source & closed-source & \textbf{84.11} & \textbf{56.16} & 60.85 & 13.79 & \textbf{72.48} & \textbf{34.98} \\

\bottomrule
\end{tabular}
\caption{MM-Vet v2 evaluation results on various VLMs regarding each prompt-group and the overall average score. For each column, the highest figure is shown \textbf{Bold}. Among the open-weight models, the first, the second, and the third highest figures are highlighted by \colorbox{YellowGreen}{green}, \colorbox{Yellow}{yellow} and \colorbox{SkyBlue}{blue} backgrounds, respectively. All the numbers are presented in \% and the full score is 100\%. The conditions in the "Statistically Significant" row are thresholds for an accuracy score to be statistically significant with p\_value$<$0.05 in its respective column.}
\label{tab:result_overall}
\end{table*}

\normalsize % Set font size to normal
%%%%%%%%%%%%%%%%%%%%%%%%%%%%%%%%%%%%%%%%%%%%%%%%

\subsection{Main Results}
\Cref{tab:result_overall} presents the evaluation results of various VLMs on both the RGB-Txt and RGB-Th-Txt prompt groups, along with their overall performance.

In the RGB-Th-Txt prompt group, Qwen2.5-VL-7B-Instruct significantly outperforms other VLMs, including the closed-source GPT4o, achieving a QAcc score of 67.26\% and an SAcc score of 17.24\%. Additionally, Mini-CPM-V-2.6, InterVL2.5-8B, and Llava-OneVision-8B-ov emerge as top-performing open-weight models.

Conversely, in the RGB-Txt prompt group, GPT4o leads the performance rankings. However, Qwen2-VL-7B-Instruct, MiniCPM-o-2.6, and Qwen2.5-VL-3B-Instruct trail behind by 3\% in QAcc and 10\% in SAcc. This suggests GPT4o exhibits stronger robustness in this prompt setting. Notably, Qwen2.5-VL-3B-Instruct surpasses many larger models, including Qwen2.5-VL-7B-Instruct, which ranks in the lower half of this prompt group, in contrast to its superior performance in the RGB-Th-Txt setting.

Overall, GPT4o ranks first, with MiniCPM-o-2.6 and Qwen2-VL-7B-Instruct emerging as strong contenders in QAcc and SAcc, respectively. Meanwhile, Chameleon-7B demonstrates the lowest performance across all prompt groups and accuracy metrics, primarily due to its poor instruction-following capability.

\begin{figure*}[h]
  \centering
  \includegraphics[height=9.6cm]{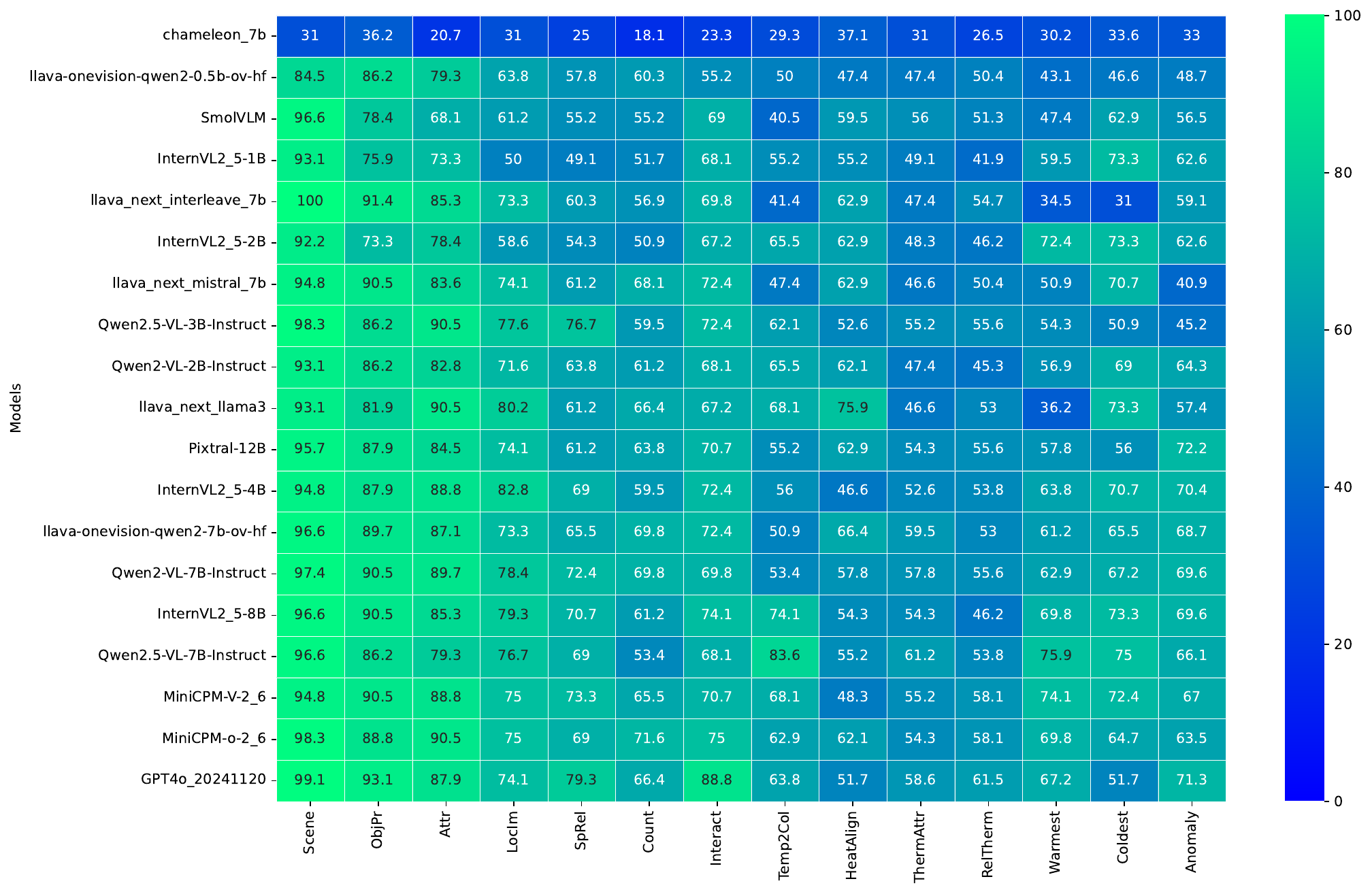}
  \caption{Illustration of models' QAcc scores across all skill dimensions, with 50\% being the random baseline performance, and where green represent higher values.}
  \label{fig:QAcc_results}
\end{figure*}

\begin{figure*}[h]
  \centering
  \includegraphics[height=9.6cm]{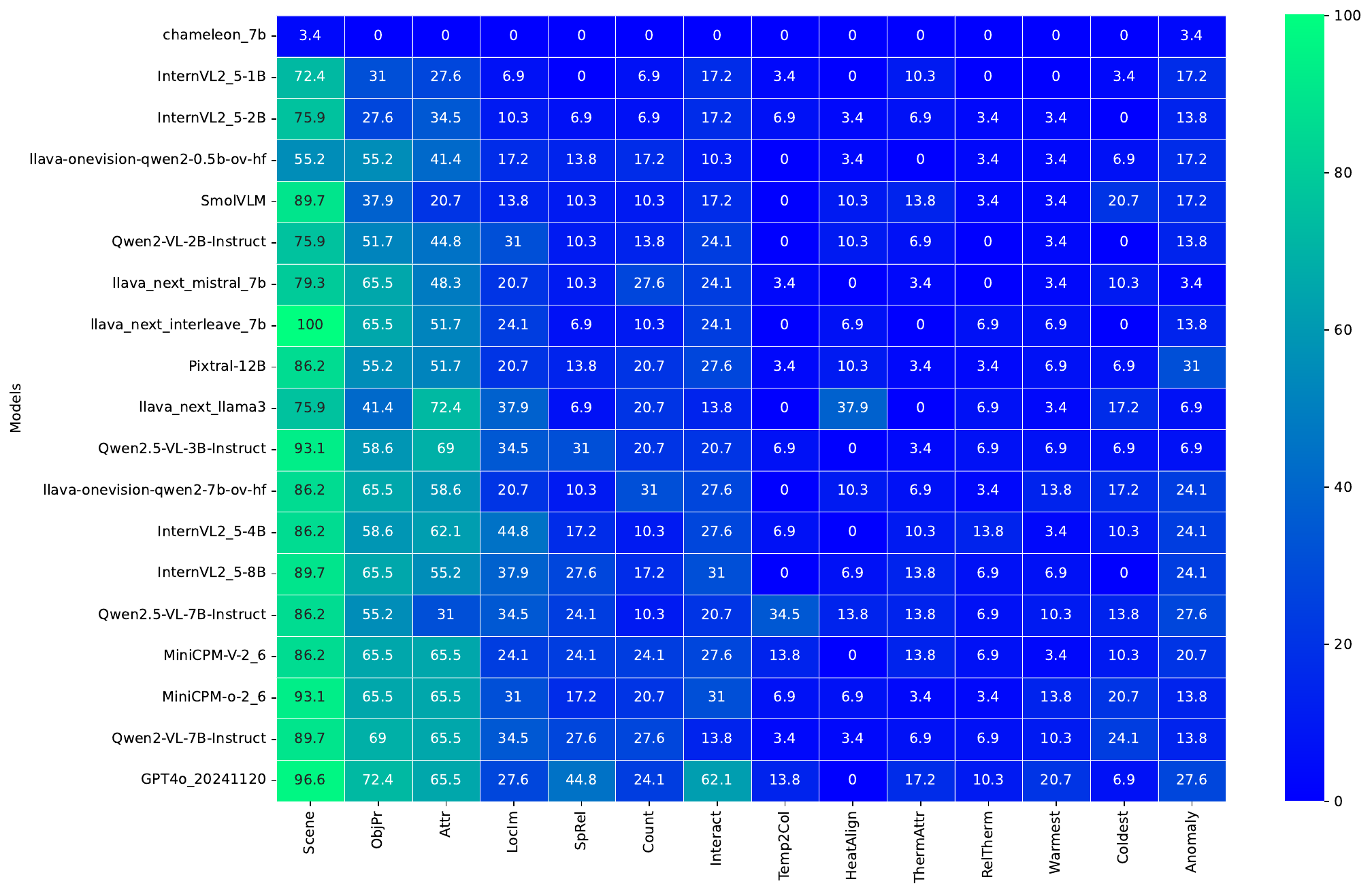}
  \caption{Illustration of models' SAcc scores across all skill dimensions, with 6.25\% being the random baseline performance, and where green represent higher values.}
  \label{fig:SAcc_results}
\end{figure*}

To further illustrate model performance across different evaluation dimensions, Figures \ref{fig:QAcc_results} and \ref{fig:SAcc_results} provide visual representations of QAcc and SAcc scores, where higher accuracy is indicated in green.

%%%%%%%%%%%%%%%%%%%%%%%%%%%%%%%%%%%%%%%%%%%%%%%%
%%%%%%%%%%%%%%%%%%%%%%%%%%%%%%%%%%%%%%%%%%%%%%%%

%%%%%%%%%%%%%%%%%%%%%%%%%%%%%%%%%%%%%%%%%%%%%%%%

\subsection{Observations}
Through an extensive evaluation of VLMs on RGB-Th-Bench, we derive key insights that can inform future research.

\textbf{The disparity in performance between RGB-Txt and RGB-Th-Txt underscores the increased complexity of comprehending RGB-Thermal data.} All VLMs perform worse on the latter prompt, which we attribute to two primary factors:
\begin{itemize}
    \item First, success in RGB-Thermal-specific skills depends on strong performance in core RGB-based skills. For instance, as illustrated in \Cref{fig:benchmark_samples}, answering "Instance Thermal Attribute" questions requires proficiency in "Detailed Object Presence" and "Instance Attributes" (\eg color). Similarly, "RGB-Thermal-Heatmap Alignment" questions necessitate strong performance in "Detailed Object Presence," "Instance Attributes" (\eg color), and "Instance Location w.r.t. Image" (\eg shift or zoom). This suggests that RGB-Thermal performance is upper-bounded by RGB-only comprehension skills.
    
    \item Second, as discussed in \Cref{sec:2_related_work}, there is currently no large-scale, expert-annotated thermal-caption dataset available for pre-training or fine-tuning. While some thermal-caption samples may exist in general image-caption datasets, their scarcity likely limits model capabilities in thermal understanding.
\end{itemize}

\noindent \textbf{Performance varies significantly across different evaluation dimensions.} This can be attributed to two key reasons:
\begin{itemize}
    \item First, the complexity of evaluation dimensions varies substantially. Certain tasks demand more sophisticated reasoning than others. For example, as shown in \Cref{fig:benchmark_samples}, all three VLMs "Pass" the SAcc criteria for "Scene Understanding" and "Detailed Object Presence," yet they "Fail" for "Instance Spatial Relation" and "Instance Interaction." A qualitative comparison reveals that the latter two require deeper contextual comprehension and reasoning.
    
    \item Second, the distribution of training data across different skills plays a crucial role. Large-scale pre-training datasets such as LAION 400M~\cite{laion400m_schuhmann2021} primarily consist of image-caption pairs, which align well with skills like "Scene Understanding," "Detailed Object Presence," and "Instance Attributes." Consequently, models demonstrate stronger performance in these dimensions across both QAcc and SAcc metrics.
\end{itemize}

\noindent \textbf{SAcc serves as a stricter and more informative metric.} For instance, as illustrated in \Cref{fig:benchmark_samples}, three of the four questions from "Temperature to Color Mapping Understanding", which are about a color sequence, are negative questions to each other—meaning a correct "Yes" response to one question necessitates a "No" response to the others. However, both GPT4o and MiniCPM-o-2.6 failed in this category, highlighting the value of SAcc in capturing comprehensive understanding.

\textbf{Most VLMs adhere to the simple "Yes/No" response format, with a few exceptions.} Notably, Chameleon~\cite{chameleon2024}, Llava-OneVision-1B-ov~\cite{tested_models_2024_llavaonevision}, and SmolVLM~\cite{tested_models_2025_smolvlm} failed to comply with this instruction 712, 6, and 2 times, respectively. Additionally, GPT4o applied content moderation by refusing to answer 28 questions. In all such cases, we classified them as "Fail." A comparison with prior benchmarks, such as MME~\cite{visling_mme_Fu2023} and MMBench~\cite{visling_MMBench_Liu2023}, which reported significant instruction-following issues, suggests that recent VLMs exhibit improved alignment in this regard.

%% file: sec/5_conclusion.tex
\section{Conclusion}
\label{sec:5_conclusion}
In this work, we introduced \textbf{RGB-Th-Bench}, the first comprehensive visual-thermal evaluation benchmark for Vision-Language Models (VLMs). RGB-Th-Bench is distinguished by its rigorous design, including its distinct purpose, high-quality data source, dense annotation coverage, and strict evaluation metrics. Our benchmark provides a robust and strict metric to assess the capabilities of VLMs in handling RGB-thermal paired data, addressing a critical gap in multimodal research.

We conducted extensive evaluations of various VLMs on RGB-Th-Bench, analyzing their performance to offer insights into different model selections. Our findings reveal that even the best-performing models, Qwen2.5-VL~\cite{tested_models_2025_Qwen25VL} and GPT4o~\cite{openai2024gpt4technicalreport}, achieve less than 18\% on our stringent SAcc metric (random baseline: 6.25\%) for RGB-Thermal-Text prompt group. Given the potential of VLMs as multimodal agents, a fundamental requirement is the ability to process diverse internal documents and generate accurate responses. The suboptimal performance observed highlights the significant progress still needed before VLMs can be effectively deployed for RGB-thermal applications.

Despite its contributions, this work has the following limitations:
\begin{itemize}
    \item Limited Dataset Size: Acquiring and annotating high-quality thermal data is costly, particularly with the detailed annotations and comprehensive skill sets required in RGB-Th-Bench.

    \item Restricted Response Format: The benchmark employs Yes/No responses to simplify accuracy measurement, eliminating the need for prompt engineering or human/LLM-based evaluations. However, this constraint limits expressive open-ended evaluations.
\end{itemize}

We hope this work provides valuable insights and serves as a foundation to guide future advancements in vision-language multimodal research, particularly in the challenging domain of RGB-thermal understanding.

%% file: sec/X_suppl.tex
\clearpage
\setcounter{page}{1}
\maketitlesupplementary

\section{Guidelines for Evaluation Dimensions}
\label{sec:guidelines}

\subsection{Single RGB \& Text}
\label{sec:single_rgb_text_guidelines}
\begin{itemize}
    \item Scene Understanding (Scene): Each question is about the major content of the image, in a way that one should be able to answer the questions by having a glimpse over the whole image, and not having to directly look at individual objects or people in detail. The questions are not related to individual objects or text in the image, but to the overall theme of the picture.
    
    \item Detailed Object Presence (ObjPr): Each question is about the presence of an object in the scene. The object can be anything with contained shape (chair, car, text) or unlimited (sky, sea, wall, floor), except text or numbers. One should be able to answer the questions by mainly focusing on the presence of the objects, rather than other aspects, such as color, shape, spatial location or its identity.
    
    \item Instance Attributes (Attr): Each question is about some visual attributes of a specific present object such as color, material, shape or fine-grained type. One should be able to answer the questions by looking at the visual appearance of the mentioned object, without considering other aspects, such as spatial location or its identity.
    
    \item Instance Location w.r.t. Image (LocIm): Each question is about location of a specific present object. The location is with respect to the image (top, bottom, left, right, center) and not relative to other objects.
    
    \item Instance Spatial Relation (SpRel): Each question is about the spatial relation between two present objects in the image. The location is relative to each other and in 3D space (up/down , left/right, back/front).
    
    \item Instance Counting (Count): Each question is about the number of appearance of a present object type. The minimum count is one.
    
    \item Instance Interaction (Interact): Each question is about the non-spatial relations and connections between two present objects. One should be able to answer the questions by finding the two mentioned objects, and slightly reason over the image to understand their interaction.
    
    % \item Numerical Text Reference Understanding (NumRef): Questions are about presence, value, metric, location, and meaning of numbers in the image, which might represent measurements, dates, quantities, or other contextual details. An example format of the questions could be: “Is this correct: there is the exact number [number] at [its location] specifically referring/showing … ?”
    
    % \item Non-Numerical Text Reference Understanding (TextRef): Questions are about the presence, meaning, purpose, and locations of non-numerical text within the image. The text could include labels, instructions, warnings, or any descriptive content. An example format of the questions could be: “Is this correct: there is the exact text [text] at [its location] specifically referring/showing … ?”

\end{itemize}
\subsection{RGB-Thermal pair \& Text}
\label{sec:rgb_thermal_text_guidelines}
\begin{itemize}
    \item Temperature to Color Mapping Understanding (Temp2Col): One or two question is dedicated to the presence of any color to temperature scale bar, and its information like the temperature unit (F or C), temperature range, etc. Other questions are about the mapping of color to temperature. One example of a question form could be “Is the the following sequence of colors correctly represent temperature distribution from low to high: [sequence of colors (at least 3 colors)] ?”
    
    \item RGB-Thermal-Heatmap Alignment (HeatAlign): Questions are either about the difference between the thermal image viewpoint and the RGB image viewpoint, or about the alignment of the heatmap on the black edges of the objects (if the thermal image has an overlay edge filter). Some example forms are:
    \begin{itemize}
        \item "Is the thermal image viewpoint exactly matching the RGB image viewpoint without any shift or zoom?",
        \item "Is the thermal image viewpoint exactly matching the RGB image viewpoint without any shift or zoom?",
        \item "Is the thermal image slightly shifted to the right/left compared to the RGB image viewpoint?"
        \item "Is the thermal image a zoomed-in crop from the center/top-left/bottom/... viewpoint of the RGB image?"
        \item "Should the heatmap be shifted down/right/left/up or zoom out/in to correctly match the black edges of the objects?"
    \end{itemize}
    
    \item Instance Thermal Attribute (ThermAttr): Each question is about some thermal attributes such as temperature intensity or density of a specific present object. One should be able to answer the questions by finding the object in both RGB and thermal images and carefully look at the visual appearance of the object in the thermal image, but does not have to consider information of other aspects, such as spatial location or its identity.
    
    \item Relative Thermal Attribute (RelTherm): Each questions is about the relative difference between temperature intensity or density of two present objects. One should be able to answer the questions by finding the two objects in both RGB and thermal images and carefully look at the visual appearance of the object in the thermal image, but does not have to consider information of other aspects, such as spatial location or its identity.
    
    \item Warmest Areas Detection (Warmest): Each question is about if a specific present area is the warmest area.
    
    \item Coldest Areas Detection (Coldest): Each question is about if a specific present area is the coldest area.
    
    \item Risk and Anomaly Detection (Anomaly): Each question is about if the thermal state of a specific present object may be considered as a risk or anomaly based on the context of the RGB and thermal images and the nature and purpose of the object. Risks and anomalies are reasonable and may be related to any poor insulation, gaps and cracks, leakage, overheating, electrical issue, fire issue, or any other issue based on the context and the function of the objects.
\end{itemize}

%% file: main.bbl
\begin{thebibliography}{57}
\providecommand{\natexlab}[1]{#1}
\providecommand{\url}[1]{\texttt{#1}}
\expandafter\ifx\csname urlstyle\endcsname\relax
  \providecommand{\doi}[1]{doi: #1}\else
  \providecommand{\doi}{doi: \begingroup \urlstyle{rm}\Url}\fi

\bibitem[FLI()]{FLIR_ONE_Edge_Pro}
{FLIR} {ONE} {Edge} {Pro} {\textbar} {Teledyne} {FLIR}.

\bibitem[Ope()]{OpenBMB-MiniCPMo-2025}
{{OpenBMB}}/{{MiniCPM-o}}.

\bibitem[cla()]{claude_sonnet}
Introducing {Claude} 3.5 {Sonnet}.

\bibitem[fli()]{flir}
{FREE} - {FLIR} {Thermal} {Dataset} for {Algorithm} {Training} {\textbar} {Teledyne} {FLIR}.

\bibitem[tno(2014)]{tno}
{TNO} {Image} {Fusion} {Dataset}, 2014.

\bibitem[Agrawal et~al.(2024)Agrawal, Antoniak, Hanna, Bout, Chaplot, Chudnovsky, Costa, Monicault, Garg, Gervet, Ghosh, Héliou, Jacob, Jiang, Khandelwal, Lacroix, Lample, Casas, Lavril, Scao, Lo, Marshall, Martin, Mensch, Muddireddy, Nemychnikova, Pellat, Platen, Raghuraman, Rozière, Sablayrolles, Saulnier, Sauvestre, Shang, Soletskyi, Stewart, Stock, Studnia, Subramanian, Vaze, Wang, and Yang]{tested_models-2024-pixtral12b}
Pravesh Agrawal, Szymon Antoniak, Emma~Bou Hanna, Baptiste Bout, Devendra Chaplot, Jessica Chudnovsky, Diogo Costa, Baudouin~De Monicault, Saurabh Garg, Theophile Gervet, Soham Ghosh, Amélie Héliou, Paul Jacob, Albert~Q. Jiang, Kartik Khandelwal, Timothée Lacroix, Guillaume Lample, Diego~Las Casas, Thibaut Lavril, Teven~Le Scao, Andy Lo, William Marshall, Louis Martin, Arthur Mensch, Pavankumar Muddireddy, Valera Nemychnikova, Marie Pellat, Patrick~Von Platen, Nikhil Raghuraman, Baptiste Rozière, Alexandre Sablayrolles, Lucile Saulnier, Romain Sauvestre, Wendy Shang, Roman Soletskyi, Lawrence Stewart, Pierre Stock, Joachim Studnia, Sandeep Subramanian, Sagar Vaze, Thomas Wang, and Sophia Yang.
\newblock Pixtral 12b, 2024.

\bibitem[Bai et~al.(2023{\natexlab{a}})Bai, Bai, Chu, Cui, Dang, Deng, Fan, Ge, Han, Huang, et~al.]{qwen_bai2023}
Jinze Bai, Shuai Bai, Yunfei Chu, Zeyu Cui, Kai Dang, Xiaodong Deng, Yang Fan, Wenbin Ge, Yu Han, Fei Huang, et~al.
\newblock Qwen technical report.
\newblock \emph{arXiv preprint arXiv:2309.16609}, 2023{\natexlab{a}}.

\bibitem[Bai et~al.(2023{\natexlab{b}})Bai, Bai, Yang, Wang, Tan, Wang, Lin, Zhou, and Zhou]{qwenvl_bai2023}
Jinze Bai, Shuai Bai, Shusheng Yang, Shijie Wang, Sinan Tan, Peng Wang, Junyang Lin, Chang Zhou, and Jingren Zhou.
\newblock Qwen-vl: A versatile vision-language model for understanding, localization, text reading, and beyond, 2023{\natexlab{b}}.

\bibitem[Barrera et~al.(2013)Barrera, Lumbreras, and Sappa]{cvc15}
Fernando Barrera, Felipe Lumbreras, and Angel~D. Sappa.
\newblock Multispectral piecewise planar stereo using manhattan-world assumption.
\newblock \emph{Pattern Recognition Letters}, 34\penalty0 (1):\penalty0 52--61, 2013.
\newblock Extracting Semantics from Multi-Spectrum Video.

\bibitem[Barrera~Campo et~al.(2012)Barrera~Campo, Lumbreras~Ruiz, and Sappa]{cvc13}
Fernando Barrera~Campo, Felipe Lumbreras~Ruiz, and Angel~Domingo Sappa.
\newblock Multimodal stereo vision system: 3d data extraction and algorithm evaluation.
\newblock \emph{IEEE Journal of Selected Topics in Signal Processing}, 6\penalty0 (5):\penalty0 437--446, 2012.

\bibitem[Cai et~al.(2024)Cai, Cao, Chen, Chen, Chen, Chen, Chen, Chen, Chen, Chu, Dong, Duan, Fan, Fei, Gao, Ge, Gu, Gu, Gui, Guo, Guo, He, Hu, Huang, Jiang, Jiao, Jin, Lei, Li, Li, Li, Li, Li, Li, Liu, Liu, Hong, Liu, Liu, Liu, Lv, Lv, Lv, Ma, Ma, Ma, Ning, Ouyang, Qiu, Qu, Shang, Shao, Song, Song, Sui, Sun, Sun, Tang, Wang, Wang, Wang, Wang, Wang, Wang, Wang, Wei, Weng, Wu, Xiong, Xu, Xu, Yan, Yan, Yang, Ye, Ying, Yu, Yu, Zang, Zhang, Zhang, Zhang, Zhang, Zhang, Zhang, Zhang, Zhang, Zhang, Zhang, Zhang, Zhao, Zhao, Zhao, Zhou, Zhou, Zhuo, Zou, Qiu, Qiao, and Lin]{cai2024internlm2}
Zheng Cai, Maosong Cao, Haojiong Chen, Kai Chen, Keyu Chen, Xin Chen, Xun Chen, Zehui Chen, Zhi Chen, Pei Chu, Xiaoyi Dong, Haodong Duan, Qi Fan, Zhaoye Fei, Yang Gao, Jiaye Ge, Chenya Gu, Yuzhe Gu, Tao Gui, Aijia Guo, Qipeng Guo, Conghui He, Yingfan Hu, Ting Huang, Tao Jiang, Penglong Jiao, Zhenjiang Jin, Zhikai Lei, Jiaxing Li, Jingwen Li, Linyang Li, Shuaibin Li, Wei Li, Yining Li, Hongwei Liu, Jiangning Liu, Jiawei Hong, Kaiwen Liu, Kuikun Liu, Xiaoran Liu, Chengqi Lv, Haijun Lv, Kai Lv, Li Ma, Runyuan Ma, Zerun Ma, Wenchang Ning, Linke Ouyang, Jiantao Qiu, Yuan Qu, Fukai Shang, Yunfan Shao, Demin Song, Zifan Song, Zhihao Sui, Peng Sun, Yu Sun, Huanze Tang, Bin Wang, Guoteng Wang, Jiaqi Wang, Jiayu Wang, Rui Wang, Yudong Wang, Ziyi Wang, Xingjian Wei, Qizhen Weng, Fan Wu, Yingtong Xiong, Chao Xu, Ruiliang Xu, Hang Yan, Yirong Yan, Xiaogui Yang, Haochen Ye, Huaiyuan Ying, Jia Yu, Jing Yu, Yuhang Zang, Chuyu Zhang, Li Zhang, Pan Zhang, Peng Zhang, Ruijie Zhang, Shuo Zhang, Songyang Zhang, Wenjian Zhang,
  Wenwei Zhang, Xingcheng Zhang, Xinyue Zhang, Hui Zhao, Qian Zhao, Xiaomeng Zhao, Fengzhe Zhou, Zaida Zhou, Jingming Zhuo, Yicheng Zou, Xipeng Qiu, Yu Qiao, and Dahua Lin.
\newblock Internlm2 technical report, 2024.

\bibitem[Chen et~al.(2024)Chen, Wang, Tian, Ye, Gao, Cui, Tong, Hu, Luo, Ma, et~al.]{chen2024internvl1.5}
Zhe Chen, Weiyun Wang, Hao Tian, Shenglong Ye, Zhangwei Gao, Erfei Cui, Wenwen Tong, Kongzhi Hu, Jiapeng Luo, Zheng Ma, et~al.
\newblock How far are we to gpt-4v? closing the gap to commercial multimodal models with open-source suites.
\newblock \emph{arXiv preprint arXiv:2404.16821}, 2024.

\bibitem[Chen et~al.(2025)Chen, Wang, Cao, Liu, Gao, Cui, Zhu, Ye, Tian, Liu, Gu, Wang, Li, Ren, Chen, Luo, Wang, Jiang, Wang, He, Shi, Zhang, Lv, Wang, Shao, Chu, Tu, He, Wu, Deng, Ge, Chen, Zhang, Wang, Dou, Lu, Zhu, Lu, Lin, Qiao, Dai, and Wang]{tested_models_2025_internvl2.5}
Zhe Chen, Weiyun Wang, Yue Cao, Yangzhou Liu, Zhangwei Gao, Erfei Cui, Jinguo Zhu, Shenglong Ye, Hao Tian, Zhaoyang Liu, Lixin Gu, Xuehui Wang, Qingyun Li, Yimin Ren, Zixuan Chen, Jiapeng Luo, Jiahao Wang, Tan Jiang, Bo Wang, Conghui He, Botian Shi, Xingcheng Zhang, Han Lv, Yi Wang, Wenqi Shao, Pei Chu, Zhongying Tu, Tong He, Zhiyong Wu, Huipeng Deng, Jiaye Ge, Kai Chen, Kaipeng Zhang, Limin Wang, Min Dou, Lewei Lu, Xizhou Zhu, Tong Lu, Dahua Lin, Yu Qiao, Jifeng Dai, and Wenhai Wang.
\newblock Expanding performance boundaries of open-source multimodal models with model, data, and test-time scaling, 2025.

\bibitem[Contributors(2023)]{2023opencompass}
OpenCompass Contributors.
\newblock Opencompass: A universal evaluation platform for foundation models.
\newblock \url{https://github.com/open-compass/opencompass}, 2023.

\bibitem[Davis and Keck(2005)]{OTCBVS}
James~W. Davis and Mark~A. Keck.
\newblock A two-stage template approach to person detection in thermal imagery.
\newblock In \emph{2005 Seventh IEEE Workshops on Applications of Computer Vision (WACV/MOTION'05) - Volume 1}, pages 364--369, 2005.

\bibitem[Duan et~al.(2024)Duan, Yang, Qiao, Fang, Chen, Liu, Dong, Zang, Zhang, Wang, et~al.]{duan2024vlmevalkit}
Haodong Duan, Junming Yang, Yuxuan Qiao, Xinyu Fang, Lin Chen, Yuan Liu, Xiaoyi Dong, Yuhang Zang, Pan Zhang, Jiaqi Wang, et~al.
\newblock Vlmevalkit: An open-source toolkit for evaluating large multi-modality models.
\newblock In \emph{Proceedings of the 32nd ACM International Conference on Multimedia}, pages 11198--11201, 2024.

\bibitem[Dubey et~al.(2024)Dubey, Jauhri, Pandey, Kadian, Al-Dahle, Letman, Mathur, Schelten, Yang, Fan, et~al.]{dubey2024llama3}
Abhimanyu Dubey, Abhinav Jauhri, Abhinav Pandey, Abhishek Kadian, Ahmad Al-Dahle, Aiesha Letman, Akhil Mathur, Alan Schelten, Amy Yang, Angela Fan, et~al.
\newblock The llama 3 herd of models.
\newblock \emph{arXiv preprint arXiv:2407.21783}, 2024.

\bibitem[Fu et~al.(2023)Fu, Chen, Shen, Qin, Zhang, Lin, Qiu, Lin, Yang, Zheng, Li, Sun, and Ji]{visling_mme_Fu2023}
Chaoyou Fu, Peixian Chen, Yunhang Shen, Yulei Qin, Mengdan Zhang, Xu Lin, Zhenyu Qiu, Wei Lin, Jinrui Yang, Xiawu Zheng, Ke Li, Xing Sun, and Rongrong Ji.
\newblock Mme: A comprehensive evaluation benchmark for multimodal large language models.
\newblock \emph{ArXiv}, abs/2306.13394, 2023.

\bibitem[Fu et~al.(2019)Fu, Qin, Luo, Wu, and Sun]{27UAV}
Zhitao Fu, Qianqing Qin, Bin Luo, Chun Wu, and Hong Sun.
\newblock A local feature descriptor based on combination of structure and texture information for multispectral image matching.
\newblock \emph{IEEE Geoscience and Remote Sensing Letters}, 16\penalty0 (1):\penalty0 100--104, 2019.

\bibitem[González et~al.(2016)González, Fang, Socarras, Serrat, Vázquez, Xu, and López]{cvc14}
Alejandro González, Zhijie Fang, Yainuvis Socarras, Joan Serrat, David Vázquez, Jiaolong Xu, and Antonio~M. López.
\newblock Pedestrian detection at day/night time with visible and fir cameras: A comparison.
\newblock \emph{Sensors}, 16\penalty0 (6), 2016.

\bibitem[Guan et~al.(2024)Guan, Liu, Wu, Xian, Li, Liu, Wang, Chen, Huang, Yacoob, et~al.]{guan2024hallusionbench}
Tianrui Guan, Fuxiao Liu, Xiyang Wu, Ruiqi Xian, Zongxia Li, Xiaoyu Liu, Xijun Wang, Lichang Chen, Furong Huang, Yaser Yacoob, et~al.
\newblock Hallusionbench: an advanced diagnostic suite for entangled language hallucination and visual illusion in large vision-language models.
\newblock In \emph{Proceedings of the IEEE/CVF Conference on Computer Vision and Pattern Recognition}, pages 14375--14385, 2024.

\bibitem[Hwang et~al.(2015)Hwang, Park, Kim, Choi, and Kweon]{dataset_kaist_pedestrian}
Soonmin Hwang, Jaesik Park, Namil Kim, Yukyung Choi, and In~So Kweon.
\newblock Multispectral pedestrian detection: Benchmark dataset and baseline.
\newblock In \emph{2015 IEEE Conference on Computer Vision and Pattern Recognition (CVPR)}, pages 1037--1045, 2015.

\bibitem[INO()]{ino_video_nodate}
INO.
\newblock Video {Analytics} {Dataset} - {Technologies}.

\bibitem[Jia et~al.(2021)Jia, Zhu, Li, Tang, and Zhou]{dataset_llvip_rgbthermalpair_pedestrian}
Xinyu Jia, Chuang Zhu, Minzhen Li, Wenqi Tang, and Wenli Zhou.
\newblock Llvip: A visible-infrared paired dataset for low-light vision.
\newblock In \emph{Proceedings of the IEEE/CVF International Conference on Computer Vision}, pages 3496--3504, 2021.

\bibitem[Kil et~al.(2024)Kil, Mai, Lee, Wang, Cheng, Wang, Liu, Chowdhury, and Chao]{ComparativeReasoningBenchmark}
Jihyung Kil, Zheda Mai, Justin Lee, Zihe Wang, Kerrie Cheng, Lemeng Wang, Ye Liu, Arpita Chowdhury, and Wei-Lun Chao.
\newblock Compbench: A comparative reasoning benchmark for multimodal llms.
\newblock \emph{Advances in Neural Information Processing Systems}, 2024.

\bibitem[Li et~al.(2023{\natexlab{a}})Li, Ge, Ge, Wang, Wang, Zhang, and Shan]{visling_seedbench2_li2023}
Bohao Li, Yuying Ge, Yixiao Ge, Guangzhi Wang, Rui Wang, Ruimao Zhang, and Ying Shan.
\newblock Seed-bench-2: Benchmarking multimodal large language models, 2023{\natexlab{a}}.

\bibitem[Li et~al.(2023{\natexlab{b}})Li, Wang, Wang, Ge, Ge, and Shan]{visling_seedbench_li2023}
Bohao Li, Rui Wang, Guangzhi Wang, Yuying Ge, Yixiao Ge, and Ying Shan.
\newblock Seed-bench: Benchmarking multimodal llms with generative comprehension, 2023{\natexlab{b}}.

\bibitem[Li et~al.(2024{\natexlab{a}})Li, Ge, Chen, Ge, Zhang, and Shan]{visling_seedbench2_plus_li2024}
Bohao Li, Yuying Ge, Yi Chen, Yixiao Ge, Ruimao Zhang, and Ying Shan.
\newblock Seed-bench-2-plus: Benchmarking multimodal large language models with text-rich visual comprehension, 2024{\natexlab{a}}.

\bibitem[Li et~al.(2024{\natexlab{b}})Li, Lin, Peng, Nyandwi, Jiang, Ma, Khanuja, Krishna, Neubig, and Ramanan]{naturalbench}
Baiqi Li, Zhiqiu Lin, Wenxuan Peng, Jean de~Dieu Nyandwi, Daniel Jiang, Zixian Ma, Simran Khanuja, Ranjay Krishna, Graham Neubig, and Deva Ramanan.
\newblock Naturalbench: Evaluating vision-language models on natural adversarial samples.
\newblock In \emph{The Thirty-eight Conference on Neural Information Processing Systems Datasets and Benchmarks Track}, 2024{\natexlab{b}}.

\bibitem[Li et~al.(2024{\natexlab{c}})Li, Zhang, Zhang, Guo, Zhang, Li, Zhang, Liu, and Li]{li-2024-llava-next-llama3}
Bo Li, Kaichen Zhang, Hao Zhang, Dong Guo, Renrui Zhang, Feng Li, Yuanhan Zhang, Ziwei Liu, and Chunyuan Li.
\newblock Llava-next: Stronger llms supercharge multimodal capabilities in the wild, 2024{\natexlab{c}}.

\bibitem[Li et~al.(2024{\natexlab{d}})Li, Zhang, Guo, Zhang, Li, Zhang, Zhang, Li, Liu, and Li]{li2024llavaonevision}
Bo Li, Yuanhan Zhang, Dong Guo, Renrui Zhang, Feng Li, Hao Zhang, Kaichen Zhang, Yanwei Li, Ziwei Liu, and Chunyuan Li.
\newblock Llava-onevision: Easy visual task transfer, 2024{\natexlab{d}}.

\bibitem[Li et~al.(2024{\natexlab{e}})Li, Zhang, Guo, Zhang, Li, Zhang, Zhang, Zhang, Li, Liu, and Li]{tested_models_2024_llavaonevision}
Bo Li, Yuanhan Zhang, Dong Guo, Renrui Zhang, Feng Li, Hao Zhang, Kaichen Zhang, Peiyuan Zhang, Yanwei Li, Ziwei Liu, and Chunyuan Li.
\newblock Llava-onevision: Easy visual task transfer, 2024{\natexlab{e}}.

\bibitem[Li et~al.(2024{\natexlab{f}})Li, Zhang, Zhang, Zhang, Li, Li, Ma, and Li]{tested_models_li-2024-llava-next-interleave}
Feng Li, Renrui Zhang, Hao Zhang, Yuanhan Zhang, Bo Li, Wei Li, Zejun Ma, and Chunyuan Li.
\newblock Llava-next-interleave: Tackling multi-image, video, and 3d in large multimodal models, 2024{\natexlab{f}}.

\bibitem[Li et~al.(2023{\natexlab{c}})Li, Du, Zhou, Wang, Zhao, and rong Wen]{hallucination_POPE_Li2023EvaluatingOH}
Yifan Li, Yifan Du, Kun Zhou, Jinpeng Wang, Wayne~Xin Zhao, and Ji rong Wen.
\newblock Evaluating object hallucination in large vision-language models.
\newblock In \emph{Conference on Empirical Methods in Natural Language Processing}, 2023{\natexlab{c}}.

\bibitem[Lin et~al.(2024)Lin, Li, Gao, Wu, Yan, Yang, Wang, and Shou]{lin2024videogui-bench}
Kevin~Qinghong Lin, Linjie Li, Difei Gao, Qinchen Wu, Mingyi Yan, Zhengyuan Yang, Lijuan Wang, and Mike~Zheng Shou.
\newblock Videogui: A benchmark for gui automation from instructional videos.
\newblock \emph{arXiv preprint arXiv:2406.10227}, 2024.

\bibitem[Lin et~al.(2014)Lin, Maire, Belongie, Hays, Perona, Ramanan, Doll{\'a}r, and Zitnick]{ObjDet_MSCOCO_Lin2014}
Tsung-Yi Lin, Michael Maire, Serge~J. Belongie, James Hays, Pietro Perona, Deva Ramanan, Piotr Doll{\'a}r, and C.~Lawrence Zitnick.
\newblock Microsoft coco: Common objects in context.
\newblock In \emph{European Conference on Computer Vision}, 2014.

\bibitem[Liu et~al.(2024{\natexlab{a}})Liu, Li, Li, Li, Zhang, Shen, and Lee]{liu-2024-llava-next_mistral_vicuna}
Haotian Liu, Chunyuan Li, Yuheng Li, Bo Li, Yuanhan Zhang, Sheng Shen, and Yong~Jae Lee.
\newblock Llava-next: Improved reasoning, ocr, and world knowledge, 2024{\natexlab{a}}.

\bibitem[Liu et~al.(2024{\natexlab{b}})Liu, Li, Li, Li, Zhang, Shen, and Lee]{llavanext_liu2024}
Haotian Liu, Chunyuan Li, Yuheng Li, Bo Li, Yuanhan Zhang, Sheng Shen, and Yong~Jae Lee.
\newblock Llava-next: Improved reasoning, ocr, and world knowledge, 2024{\natexlab{b}}.

\bibitem[Liu et~al.(2024{\natexlab{c}})Liu, Li, Wu, and Lee]{llava_liu2024}
Haotian Liu, Chunyuan Li, Qingyang Wu, and Yong~Jae Lee.
\newblock Visual instruction tuning.
\newblock \emph{Advances in neural information processing systems}, 36, 2024{\natexlab{c}}.

\bibitem[Liu et~al.(2023)Liu, Duan, Zhang, Li, Zhang, Zhao, Yuan, Wang, He, Liu, Chen, and Lin]{visling_MMBench_Liu2023}
Yuanzhan Liu, Haodong Duan, Yuanhan Zhang, Bo Li, Songyang Zhang, Wangbo Zhao, Yike Yuan, Jiaqi Wang, Conghui He, Ziwei Liu, Kai Chen, and Dahua Lin.
\newblock Mmbench: Is your multi-modal model an all-around player?
\newblock \emph{ArXiv}, abs/2307.06281, 2023.

\bibitem[Lu et~al.(2022)Lu, Mishra, Xia, Qiu, Chang, Zhu, Tafjord, Clark, and Kalyan]{visling_science_qa_Lu2022}
Pan Lu, Swaroop Mishra, Tony Xia, Liang Qiu, Kai-Wei Chang, Song-Chun Zhu, Oyvind Tafjord, Peter Clark, and A. Kalyan.
\newblock Learn to explain: Multimodal reasoning via thought chains for science question answering.
\newblock \emph{ArXiv}, abs/2209.09513, 2022.

\bibitem[Marafioti et~al.(2025)Marafioti, Noyan, Farré, Bakouch, and Cuenca]{tested_models_2025_smolvlm}
Andres Marafioti, Merve Noyan, Miquel Farré, Elie Bakouch, and Pedro Cuenca.
\newblock {SmolVLM} - small yet mighty {Vision} {Language} {Model}, 2025.

\bibitem[OpenAI(2024)]{openai2024gpt4technicalreport}
OpenAI.
\newblock Gpt-4 technical report, 2024.

\bibitem[Reid et~al.(2024)Reid, Savinov, Teplyashin, Lepikhin, Lillicrap, Alayrac, Soricut, Lazaridou, Firat, Schrittwieser, et~al.]{gemini_pro_reid2024}
Machel Reid, Nikolay Savinov, Denis Teplyashin, Dmitry Lepikhin, Timothy Lillicrap, Jean-baptiste Alayrac, Radu Soricut, Angeliki Lazaridou, Orhan Firat, Julian Schrittwieser, et~al.
\newblock Gemini 1.5: Unlocking multimodal understanding across millions of tokens of context.
\newblock \emph{arXiv preprint arXiv:2403.05530}, 2024.

\bibitem[Schuhmann et~al.(2021)Schuhmann, Vencu, Beaumont, Kaczmarczyk, Mullis, Katta, Coombes, Jitsev, and Komatsuzaki]{laion400m_schuhmann2021}
Christoph Schuhmann, Richard Vencu, Romain Beaumont, Robert Kaczmarczyk, Clayton Mullis, Aarush Katta, Theo Coombes, Jenia Jitsev, and Aran Komatsuzaki.
\newblock Laion-400m: Open dataset of clip-filtered 400 million image-text pairs.
\newblock \emph{arXiv preprint arXiv:2111.02114}, 2021.

\bibitem[Si et~al.(2024)Si, Zhang, Yang, Liu, and Yang]{si2024design2code}
Chenglei Si, Yanzhe Zhang, Zhengyuan Yang, Ruibo Liu, and Diyi Yang.
\newblock Design2code: How far are we from automating front-end engineering?
\newblock \emph{arXiv preprint arXiv:2403.03163}, 2024.

\bibitem[Team(2024)]{chameleon2024}
Chameleon Team.
\newblock Chameleon: Mixed-modal early-fusion foundation models, 2024.

\bibitem[Team()]{tested_models_2025_Qwen25VL}
Qwen Team.
\newblock Qwen2.5 {{VL}}! {{Qwen2}}.5 {{VL}}! {{Qwen2}}.5 {{VL}}!

\bibitem[Touvron et~al.(2023)Touvron, Lavril, Izacard, Martinet, Lachaux, Lacroix, Rozi{\`e}re, Goyal, Hambro, Azhar, Rodriguez, Joulin, Grave, and Lample]{LLaMA_Touvron2023}
Hugo Touvron, Thibaut Lavril, Gautier Izacard, Xavier Martinet, Marie-Anne Lachaux, Timoth{\'e}e Lacroix, Baptiste Rozi{\`e}re, Naman Goyal, Eric Hambro, Faisal Azhar, Aurelien Rodriguez, Armand Joulin, Edouard Grave, and Guillaume Lample.
\newblock Llama: Open and efficient foundation language models.
\newblock \emph{ArXiv}, abs/2302.13971, 2023.

\bibitem[Wagstaff(2012)]{wagstaff2012ml_matters}
Kiri Wagstaff.
\newblock Machine learning that matters.
\newblock \emph{arXiv preprint arXiv:1206.4656}, 2012.

\bibitem[Wang et~al.(2025)Wang, Lee, Menghini, Mols, Doughty, Khoja, Lynch, Hendryx, Yue, and Hendrycks]{wang2025enigmaevalbenchmarklongmultimodal}
Clinton~J. Wang, Dean Lee, Cristina Menghini, Johannes Mols, Jack Doughty, Adam Khoja, Jayson Lynch, Sean Hendryx, Summer Yue, and Dan Hendrycks.
\newblock Enigmaeval: A benchmark of long multimodal reasoning challenges, 2025.

\bibitem[Wang et~al.(2024)Wang, Bai, Tan, Wang, Fan, Bai, Chen, Liu, Wang, Ge, Fan, Dang, Du, Ren, Men, Liu, Zhou, Zhou, and Lin]{Qwen2VL}
Peng Wang, Shuai Bai, Sinan Tan, Shijie Wang, Zhihao Fan, Jinze Bai, Keqin Chen, Xuejing Liu, Jialin Wang, Wenbin Ge, Yang Fan, Kai Dang, Mengfei Du, Xuancheng Ren, Rui Men, Dayiheng Liu, Chang Zhou, Jingren Zhou, and Junyang Lin.
\newblock Qwen2-vl: Enhancing vision-language model's perception of the world at any resolution.
\newblock \emph{arXiv preprint arXiv:2409.12191}, 2024.

\bibitem[Yao et~al.(2024)Yao, Yu, Zhang, Wang, Cui, Zhu, Cai, Li, Zhao, He, et~al.]{2024-minicpm-v}
Yuan Yao, Tianyu Yu, Ao Zhang, Chongyi Wang, Junbo Cui, Hongji Zhu, Tianchi Cai, Haoyu Li, Weilin Zhao, Zhihui He, et~al.
\newblock Minicpm-v: A gpt-4v level mllm on your phone.
\newblock \emph{arXiv preprint arXiv:2408.01800}, 2024.

\bibitem[Yu et~al.(2023)Yu, Yang, Li, Wang, Lin, Liu, Wang, and Wang]{visling_mmvet_yu2023}
Weihao Yu, Zhengyuan Yang, Linjie Li, Jianfeng Wang, Kevin Lin, Zicheng Liu, Xinchao Wang, and Lijuan Wang.
\newblock Mm-vet: Evaluating large multimodal models for integrated capabilities, 2023.

\bibitem[Yu et~al.(2024)Yu, Yang, Ren, Li, Wang, Lin, Lin, Liu, Wang, and Wang]{yu2024mmvet2}
Weihao Yu, Zhengyuan Yang, Linfeng Ren, Linjie Li, Jianfeng Wang, Kevin Lin, Chung-Ching Lin, Zicheng Liu, Lijuan Wang, and Xinchao Wang.
\newblock Mm-vet v2: A challenging benchmark to evaluate large multimodal models for integrated capabilities.
\newblock \emph{arXiv preprint arXiv:2408.00765}, 2024.

\bibitem[Yue et~al.(2023)Yue, Ni, Zhang, Zheng, Liu, Zhang, Stevens, Jiang, Ren, Sun, Wei, Yu, Yuan, Sun, Yin, Zheng, Yang, Liu, Huang, Sun, Su, and Chen]{visling_MMMU_Yue2023}
Xiang Yue, Yuansheng Ni, Kai Zhang, Tianyu Zheng, Ruoqi Liu, Ge Zhang, Samuel Stevens, Dongfu Jiang, Weiming Ren, Yuxuan Sun, Cong Wei, Botao Yu, Ruibin Yuan, Renliang Sun, Ming Yin, Boyuan Zheng, Zhenzhu Yang, Yibo Liu, Wenhao Huang, Huan Sun, Yu Su, and Wenhu Chen.
\newblock Mmmu: A massive multi-discipline multimodal understanding and reasoning benchmark for expert agi.
\newblock \emph{2024 IEEE/CVF Conference on Computer Vision and Pattern Recognition (CVPR)}, pages 9556--9567, 2023.

\bibitem[Zhang et~al.(2024)Zhang, Dong, Zang, Cao, Qian, Chen, Guo, Duan, Wang, Ouyang, Zhang, Zhang, Li, Gao, Sun, Zhang, Li, Li, Wang, Yan, He, Zhang, Chen, Dai, Qiao, Lin, and Wang]{zhang2024internlmxcomposer25}
Pan Zhang, Xiaoyi Dong, Yuhang Zang, Yuhang Cao, Rui Qian, Lin Chen, Qipeng Guo, Haodong Duan, Bin Wang, Linke Ouyang, Songyang Zhang, Wenwei Zhang, Yining Li, Yang Gao, Peng Sun, Xinyue Zhang, Wei Li, Jingwen Li, Wenhai Wang, Hang Yan, Conghui He, Xingcheng Zhang, Kai Chen, Jifeng Dai, Yu Qiao, Dahua Lin, and Jiaqi Wang.
\newblock Internlm-xcomposer-2.5: A versatile large vision language model supporting long-contextual input and output, 2024.

\end{thebibliography}
